\pdfoutput=1
\documentclass[11pt]{article}

\usepackage[final]{acl_single_col}
\usepackage{times}
\usepackage{latexsym}

\usepackage[T1]{fontenc}
\usepackage{amsmath}
\usepackage{amssymb}
\usepackage{enumitem}
\usepackage[utf8]{inputenc}
\usepackage{microtype}

\usepackage{inconsolata}

\usepackage{graphicx}
\usepackage{booktabs}
\usepackage{caption}
\usepackage{multirow}   % 合并行
\usepackage{array}      % 对齐方式
\usepackage{subcaption}
\usepackage{float}

\newcommand{\ours}{DialogueReason}

\title{\ours: Rule-Based RL Sparks Dialogue Reasoning in LLMs}

\author{
  Yubo Shu\textsuperscript{1}\thanks{~~Equal contribution}, Zhewei Huang\textsuperscript{1}\footnotemark[1], Xin Wu\textsuperscript{1}, Chen Hu\textsuperscript{1}, Shuchang Zhou\textsuperscript{1}\thanks{~~Corresponding author}, Daxin Jiang\textsuperscript{1} \\
  \textsuperscript{1}StepFun, China \\
  \texttt{shuchang.zhou@gmail.com} \\
  \textbf{HuggingFace:} \url{https://huggingface.co/stepfun-ai/Qwen2.5-32B-DialogueReason}
}

\begin{document}

\maketitle

% \begin{CJK}{UTF8}{gbsn}

\begin{abstract}

We propose \textbf{\ours}, a reasoning paradigm that uncovers the lost roles in monologue-style reasoning models, aiming to boost diversity and coherency of the reasoning process. Recent advances in RL-based large reasoning models have led to impressive long CoT capabilities and high performance on math and science benchmarks. However, these reasoning models rely mainly on monologue-style reasoning, which often limits reasoning diversity and coherency, frequently recycling fixed strategies or exhibiting unnecessary shifts in attention. Our work consists of an analysis of monologue reasoning patterns and the development of a dialogue-based reasoning approach. We first introduce the Compound-QA task, which concatenates multiple problems into a single prompt to assess both diversity and coherency of reasoning. Our analysis shows that Compound-QA exposes weaknesses in monologue reasoning, evidenced by both quantitative metrics and qualitative reasoning traces. Building on the analysis, we propose a dialogue-based reasoning, named \ours, structured around agents, environment, and interactions.  Using PPO with rule-based rewards, we train open-source LLMs (Qwen-QWQ and Qwen-Base) to adopt dialogue reasoning. We evaluate trained models on MATH, AIME, and GPQA datasets, showing that the dialogue reasoning model outperforms monologue models under more complex compound questions. Additionally, we discuss how dialogue-based reasoning helps enhance interpretability, facilitate more intuitive human interaction, and inspire advances in multi-agent system design.

\end{abstract}

\section{Introduction}

As an emergent capability of large language models (LLMs) with extensive parameter sizes during the reasoning process, chain-of-thought (CoT) allocates additional computational resources in a manner that promotes the gradual decomposition of tasks, a feature notably evident in context learning~\cite{wei2022chain}. Consequently, chain-of-thought offers methodologies for progressively revealing intermediate reasoning steps, thereby bolstering LLMs' proficiency in addressing complex problems by breaking them down into coherent sub-steps~\cite{feng2023towards}. As an extension, Long-CoT offers longer reasoning paths and more complex reasoning steps, thereby further enhancing the reasoning capabilities of LLMs~\cite{jaech2024openai,guo2025deepseek}. 

For large reasoning models~(LRMs), recent advancements have shown remarkable success, particularly with the integration of reinforcement learning (RL) to improve their ability to perform long-form CoT reasoning. Typically, these models structure their reasoning within a dedicated <think> block \cite{guo2025deepseek}, where the detailed monologue reasoning process unfolds, followed by an <answer> block that explicitly provides the final solution. Such LRMs demonstrate impressive capabilities in reflection, self-verification, and critical analysis, achieving state-of-the-art results in mathematics and science tasks. However, \textbf{these monologue-style LRMs exhibit low diversity and low coherency} in their reasoning processes. Low diversity occurs when models persistently apply fixed strategies across diverse problems, resulting in performance degradation when problems require different approaches. Low coherency arises from frequent shifts in attention within a single reasoning path, exemplified by repetitive hesitations such as "Wait…" or unnecessary switches to alternative ideas. Consequently, the reasoning process becomes fragmented, difficult to interpret, and often ineffective, swinging between overcommitting to a strategy and neglecting manifold possibilities \cite{wang2025thoughts,chen2024not}.

To address the limitations, \textbf{we first analyze the monologue reasoning patterns with a focus on diversity and coherency, and subsequently propose a dialogue-based reasoning pattern}. Specifically, inspired by the structure of divergent and convergent thinking, we introduce the Compound Question Answering (Compound-QA) task to systematically evaluate monologue reasoning models. By concatenating multiple independently solvable problems into a single input prompt, the Compound-QA task intrinsically evaluates a model's ability to employ diverse reasoning strategies while maintaining internal coherency across different reasoning paths. Our analysis further demonstrates that Compound-QA effectively reveals the weaknesses of current monologue reasoning models, both quantitatively through performance metrics and qualitatively through detailed examination of their reasoning traces.

Building upon these insights and drawing inspiration from multi-agent simulation, we articulate the design space for our proposed dialogue reasoning pattern. The design space is composed of three core dimensions: individual reasoning agents, the reasoning environment, and interaction settings. We leverage Proximal Policy Optimization (PPO) with rule-based reward functions to guide both the Qwen-QWQ-32B reasoning model and the base Qwen2.5-Base-32B model toward adopting the dialogue-based reasoning approach \cite{qwq32b, qwen2.5}.

To evaluate dialogue reasoning, we conduct comparative experiments based on MATH, AIME and GPQA datasets. Our results indicate that the dialogue reasoning model outperforms the monologue reasoning model, demonstrating enhanced accuracy on compound QA tasks, as well as superior diversity and coherency in the reasoning process. Furthermore, we explore broader implications and additional advantages offered by the dialogue reasoning pattern, including enhanced interpretability of the reasoning process, improved controllability and user interaction, and its potential to inspire end-to-end multi-agent systems beyond existing systems like the Manus series~\cite{openmanus2025}.

In summary, our core contributions are:
\begin{itemize} [itemsep=0pt, topsep=2pt, parsep=0pt]
\item Introducing the Compound-QA task, specifically designed to identify and highlight the diversity and coherency shortcomings of current monologue-based reasoning models.
\item Defining the dialogue-based reasoning framework inspired by the multi-agent system and effectively training monologue reasoning models to adopt dialogue reasoning via RL.
\item Demonstrating the advantages of dialogue reasoning over monologue models, while highlighting additional potential benefits and applications of dialogue reasoning.

\end{itemize}

\section{Monologue Reasoning Inspection}

In this section, we first clarify the concepts of diversity and coherency, discussing their roles and significance within reasoning processes. We then introduce the Compound Question Answering (Compound-QA) task, designed specifically to assess diversity and coherency. Finally, by conducting experiments using monologue reasoning models on Compound-QA tasks, we illustrate the limitations related to diversity and coherency.

\subsection{Diversity and Coherency in Thinking}
\begin{figure}[htbp]
\centering
\includegraphics[width=0.7\textwidth]{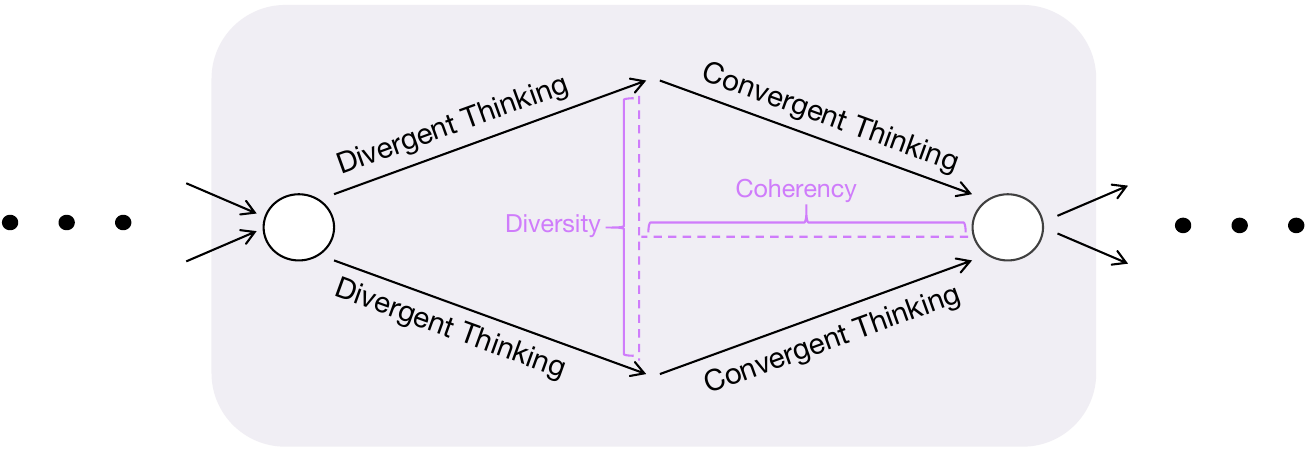}
\caption{Illustration of the reasoning process as an interplay between divergent and convergent thinking.}
\label{fig:dd_cc}
\end{figure}

The concepts of divergent and convergent thinking were originally proposed by psychologist Joy Paul Guilford during his investigation into human intelligence and creativity \cite{guilford1956structure,guilford1967nature}. According to Guilford, as illustrated in Figure~\ref{fig:dd_cc}, human reasoning operates as an iterative interplay between two fundamental cognitive modes: divergent thinking and convergent thinking.

(1) Divergent thinking refers to the cognitive process of generating distinct ideas in response to different problems. Within reasoning tasks, diversity implies the ability to shift perspectives and utilize knowledge from multiple domains rather than persistently cycling within a narrow set of ideas. High diversity increases the likelihood of discovering novel and useful solutions. Conversely, low diversity often leads to underthinking candidate perspectives and overcommitting to a fixed, possibly inadequate set of ideas. 

(2) Convergent thinking involves the cognitive process of systematically narrowing multiple possibilities down to a limited set of correct or optimal solutions. Within this process, coherency plays a crucial role by sustaining focused attention and maintaining persistent progress along a single reasoning path, enabling thorough exploitation of available information. In contrast, low coherency often results in frequent shifts between ideas, commonly reflected in repetitive "wait" and "alternative" behaviors, which disrupt the reasoning flow and increase the risk of losing track of promising solution paths \cite{wang2025thoughts}.

Considering that effective reasoning depends on a balanced interplay between divergent and convergent thinking, it is crucial for reasoning models to harmonize these two cognitive modes, avoiding conflicts, especially when attempting divergent and convergent thinking simultaneously. Instead, an optimal reasoning process typically involves an iterative cycle, alternating systematically between divergence to explore possibilities and convergence to solidify conclusions.

\subsection{Compound-QA Task}

Current quantitative metrics, such as BLEU~\cite{papineni2002bleu}, ROUGE~\cite{lin2004rouge}, and perplexity, primarily evaluate textual quality rather than the underlying reasoning processes. Consequently, these metrics are inadequate for assessing diversity and coherency of reasoning. Given the absence of established metrics, we propose an alternative evaluation approach based on the following principles:

\begin{enumerate}[label=(\arabic*), itemsep=0pt, topsep=2pt, parsep=0pt]
\item Task completion should intrinsically require both reasoning diversity and coherency.
\item Answer correctness must be objectively verifiable to avoid subjective bias.
\item The degree of requirement for diversity and coherency should be controllable for comparison.
\end{enumerate}

Based on these considerations, we construct the evaluation task by concatenating multiple individually solvable questions that the model has demonstrated the capability to handle separately. This design explicitly tests two critical aspects of reasoning:

\begin{itemize}[itemsep=0pt, topsep=2pt, parsep=0pt]
  \item \textbf{Reasoning Diversity:} When facing heterogeneous sub-problems within a single input, the model should flexibly adopt \emph{different} solution strategies rather than rigidly applying the same approach to all tasks. For example, combinatorial problems might require breadth-first search (BFS) to enumerate possible solutions, whereas geometric proofs might rely on depth-first search (DFS) for detailed deductive reasoning.
  \item \textbf{Reasoning Coherency:} Once the model chooses a particular reasoning path, it should \emph{consistently and thoroughly} explore that path to derive a robust conclusion rather than frequently switching paths and consequently losing track of the correct solution.
\end{itemize}

% These attributes align closely with J. P. Guilford's Structure-of-Intellect theory, where \emph{divergent thinking} emphasizes exploring multiple reasoning paths, while \emph{convergent thinking} stresses sustained and focused exploration of a single path to completion. 

\begin{figure}[htbp]
\centering
\includegraphics[width=0.6\textwidth]{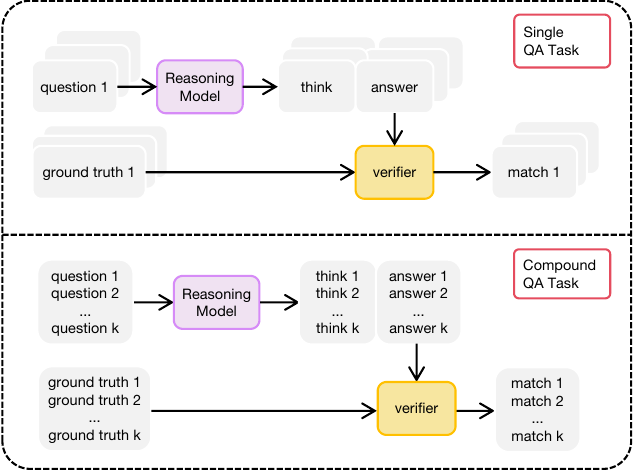}
\caption{Illustration of the Compound-QA task, which involves reasoning over multiple sub-questions concatenated into a single prompt. The figure also highlights the relationship between Compound-QA and Single-QA.}
\label{fig:compound_qa}
\end{figure}

\vspace{-5pt}
\paragraph{Compound-QA Task Definition.} We first define the original question set as $\mathcal{Q}=\{q_1,q_2,\dots,q_N\}$, where each question $q_i$ has a \emph{unique} correct answer $a_i$, and the model is assumed to be able to solve each question independently when presented individually. Figure~\ref{fig:compound_qa} illustrates the compound QA task and highlights its relationship to single QA.

Let $\oplus$ denote the concatenation operator. By selecting $k$ different questions from $\mathcal{Q}$:
\[
\mathbf{q}^{(k)}=\bigl(q_{i_1},q_{i_2},\dots,q_{i_k}\bigr) \qquad (k\le N,\; i_j\neq i_{j'})
\]
we define the resulting compound question as:
\[
\mathrm{CQ}_k \;=\; q_{i_1}\ \oplus\ q_{i_2}\ \oplus\ \dots\ \oplus\ q_{i_k},
\]
meaning that the $k$ sub-questions are concatenated into a single prompt and input to the model $\pi_\theta$ at once.

The model must produce an output:
\[
\hat{\mathbf{a}}^{(k)}=\bigl(\hat{a}_{i_1},\hat{a}_{i_2},\dots,\hat{a}_{i_k}\bigr),
\]
where each predicted answer $\hat{a}_{i_j}$ corresponds to the ground truth $a_{i_j}$.  
The overall accuracy is then defined as:
\[
\text{Acc}(\pi_\theta,\mathrm{CQ}_k)=\frac{1}{k}\sum_{j=1}^{k}\mathbb{I}\bigl[\hat{a}_{i_j}=a_{i_j}\bigr],
\]
where $\mathbb{I}[\cdot]$ denotes the indicator function. By adjusting $k$, we can obtain a \textbf{controllable difficulty curve}: as $k$ increases, the input context becomes longer, strategy-switching becomes more demanding, and attention management becomes more challenging, thus further exposing weaknesses in diversity and coherency.

\subsection{Results}

We conduct experiments using the MATH-500~\cite{hendrycks2021measuring}, GPQA Diamond~\cite{rein2024gpqa}, and AIME24 benchmarks, evaluating the LRM QWQ-32B model under its designed monologue pattern. Specifically, we measure the change in QWQ-32B's answer accuracy as the number of questions in a compound input increases. To mitigate the randomness of single-time inference, each compound question is answered 16 times independently. In our experimental setup, the number of questions per compound ranges from 1 to 10. The detailed experimental results are summarized in Table~\ref{tab:qwq_changes}. It is worth noting that the model's context length is 131,072 tokens, which is approximately 500,000 characters, and we allocated enough context length to support the experiments.

\begin{table}[htbp]
\footnotesize
  \centering
  \setlength{\tabcolsep}{4pt}
  \caption{QWQ-32B task accuracy and average CoT length as the compound factor cbK grows.}
  \label{tab:qwq_changes}
  \begin{tabular}{l|cccc|cccc|cccc}
    \toprule
          & \multicolumn{4}{c|}{\textbf{MATH-500}} 
          & \multicolumn{4}{c|}{\textbf{AIME24}} 
          & \multicolumn{4}{c}{\textbf{GPQA\_Diamond}} \\
    \cmidrule(lr){2-5}\cmidrule(lr){6-9}\cmidrule(lr){10-13}
          & \textbf{Acc} & $\Delta$Acc & \textbf{Len} & $\Delta$Len
          & \textbf{Acc} & $\Delta$Acc & \textbf{Len} & $\Delta$Len
          & \textbf{Acc} & $\Delta$Acc & \textbf{Len} & $\Delta$Len \\
    \midrule
\textbf{cbK1}  & 97.62\% & – & 10,952 & – & 74.38\% & – & 40,872 & – & 65.94\% & – & 30,987 & – \\
\textbf{cbK2}  & 95.35\% & \textcolor{red}{-2.27\%} & 14,676 & +3724 & 64.58\% & \textcolor{red}{-9.80\%} & 52,219 & +11,347 & 61.27\% & \textcolor{red}{-4.67\%} & 40,656 & +9,669 \\
\textbf{cbK3}  & 95.15\% & \textcolor{red}{-2.47\%} & 19,681 & +8,729 & 58.13\% & \textcolor{red}{-16.25\%} & 58,752 & +17,880 & 58.68\% & \textcolor{red}{-7.26\%} & 49,183 & +18,196 \\
\textbf{cbK4}  & 94.53\% & \textcolor{red}{-3.09\%} & 19,404 & +8,452 & 55.27\% & \textcolor{red}{-19.11\%} & 60,972 & +20,100 & 58.06\% & \textcolor{red}{-7.88\%} & 53,729 & +22,742 \\
\textbf{cbK5}  & 93.85\% & \textcolor{red}{-3.77\%} & 25,238 & +14,286 & 47.71\% & \textcolor{red}{-26.67\%} & 57,841 & +16,969 & 55.43\% & \textcolor{red}{-10.51\%} & 59,353 & +28,366 \\
\textbf{cbK6}  & 92.96\% & \textcolor{red}{-4.66\%} & 23,415 & +12,463 & 45.42\% & \textcolor{red}{-28.96\%} & 55,540 & +14,668 & 51.67\% & \textcolor{red}{-14.27\%} & 62,436 & +31,449 \\
\textbf{cbK7}  & 91.69\% & \textcolor{red}{-5.93\%} & 23,420 & +12,468 & 44.64\% & \textcolor{red}{-29.74\%} & 53,918 & +13,046 & 50.68\% & \textcolor{red}{-15.26\%} & 61,727 & +30,740 \\
\textbf{cbK8}  & 91.63\% & \textcolor{red}{-5.99\%} & 25,309 & +14,357 & 35.09\% & \textcolor{red}{-39.29\%} & 54,375 & +13,503 & 49.49\% & \textcolor{red}{-16.45\%} & 65,152 & +34,165 \\
\textbf{cbK9}  & 90.16\% & \textcolor{red}{-7.46\%} & 23,012 & +12,060 & 33.38\% & \textcolor{red}{-41.00\%} & 55,670 & +14,798 & 46.57\% & \textcolor{red}{-19.37\%} & 71,468 & +40,481 \\
\textbf{cbK10} & 89.63\% & \textcolor{red}{-7.99\%} & 23,332 & +12,380 & 26.67\% & \textcolor{red}{-47.71\%} & 42,191 & +1,319  & 45.66\% & \textcolor{red}{-20.28\%} & 72,651 & +41,664 \\
    \bottomrule
  \end{tabular}
\end{table}

Here, \textbf{cbK} stands for \textit{combination-K}. For example, cbK1 indicates that the compound question contains a single sub-question, corresponding to the standard single-question answering scenario, while cbK5 means that five questions are combined into one prompt. From the experimental results, we observe that as the number of questions in the compound input increases, the reasoning length generally grows across all datasets, while the accuracy declines. This trend is more pronounced on challenging AIME24 and GPQA Diamond datasets compared to MATH-500~\cite{hendrycks2021measuring}. We observed that under the cbK10 in AIME24, the model exhibited a relatively reduction in inference length. This phenomenon may be attributed to exceeding the model's capacity, potentially leading to a collapse in reasoning.

According to the design of our compound question task, the set of questions the model encounters remains unchanged. As cbK increases, namely the number of questions combined in a single input grows, which challenges the model's ability to maintain both diversity across different lines of reasoning and coherency within each individual reasoning path. The experiment shows that as cbK increases, the model's performance on compound questions significantly declines, further highlighting the limitations of current monologue-style reasoning models in handling diversity and coherency. It is important to note that although the average reasoning length allocated to each individual question may decrease as cbK rises, our intent in combining multiple single questions is to simulate the model's performance on more complex tasks. When approaching complex tasks, reasoning models tend to allocate more computational effort as task complexity increases, yet their performance on such tasks often diminishes. If we further explore more challenging compound question formats, focusing not just on direct concatenation but on tasks that require the model to actively decompose complexity, the negative impact of increased reasoning length on performance could become even more pronounced.

\subsection{Analysis}

\begin{figure}[H]
    \centering
    % 左边的大图
    \begin{subfigure}[b]{0.75\textwidth}
        \includegraphics[width=\textwidth]{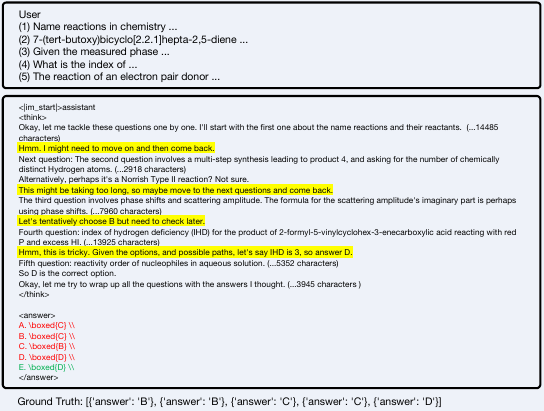} % 左边的大图
        \caption{Inability to deepen reasoning: attention deficit}
        \label{fig:analysis_monologue_left}
    \end{subfigure}
    \begin{subfigure}[b]{0.75\textwidth}
            \includegraphics[width=\textwidth]{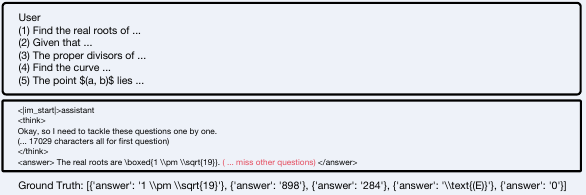} % 右上小图
            \caption{Only answering the first question}
            \label{fig:analysis_monologue_topright}
    \end{subfigure}
    \begin{subfigure}[b]{0.75\textwidth}
            \includegraphics[width=\textwidth]{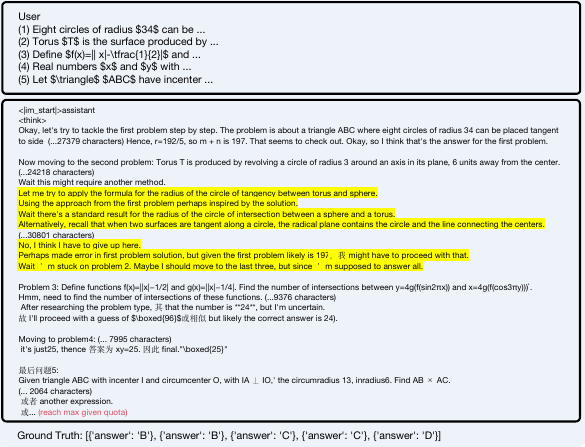} % 右下小图
            \caption{Hard to balance coherency and diversity}
            \label{fig:analysis_monologue_bottomright}
    \end{subfigure}
    \caption{Three representative failure made by monologue reasoning models when tackling compound questions.}
    \label{fig:analysis_monologue}
\end{figure}

\paragraph{Inability to deepen reasoning: attention deficit.} As shown in Figure~\ref{fig:analysis_monologue_left}, when the model receives independent chemistry and physics questions at once, it attempts to reason through each but fails to conduct any deep analysis for any individual problem. It oscillates between the chemical reactions of the first question, the hydrogen atom distinctions in the second, and the scattering amplitude calculations of the third, repeatedly shifting attention across different questions, exhibiting signs of attention deficit. Each question only receives fragmented and shallow processing. As a result, the reasoning becomes lengthy yet inefficient: although the model consumes a large number of tokens, it still fails to produce any correct answers. This case exemplifies a major failure pattern caused by low coherency in monologue reasoning: the model lacks the ability to persistently and deeply explore a single problem, making it incapable of maintaining logical depth when handling multiple concurrent tasks.
% \vspace{-5pt}

\paragraph{Only answering the first question.} As shown in Figure~\ref{fig:analysis_monologue_topright}, after receiving different sub-questions, the model only engages in chain-of-thought reasoning for the first question and provides an answer solely for it, completely ignoring the remaining four questions. This behavior highlights a limitation in the model's reasoning diversity: when facing multiple tasks, the model fails to allocate independent reasoning paths to each question, focusing all its attention on the first one. Such a single-path thinking pattern violates the natural requirement for diverse exploration in multi-question settings and severely limits the model's capacity to process complex or batch tasks in practical applications.
% \vspace{-5pt}

\paragraph{Hard to balance coherency and diversity.} As illustrated in Figure~\ref{fig:analysis_monologue_bottomright}, once the model deeply reasons through the first question (e.g., producing an answer of $r=192/5$), it attempts to reuse the same reasoning strategy for subsequent questions. The model remains trapped within the initial logical framework and struggles to adapt its reasoning to new problems. Consequently, the third, fourth, and fifth questions fail to be properly addressed, as the model's reasoning budget is already consumed by the earlier tasks. This demonstrates the rigidity of thinking inherent in monologue reasoning: early reasoning paths constrain subsequent thought processes, and the model lacks the ability to dynamically adjust or introduce new strategies for new questions. Ultimately, it fails to complete all tasks within the given reasoning budget.

\section{\ours}
\vspace{0pt}
\begin{figure}[htbp]
\centering
\includegraphics[width=0.95\textwidth]{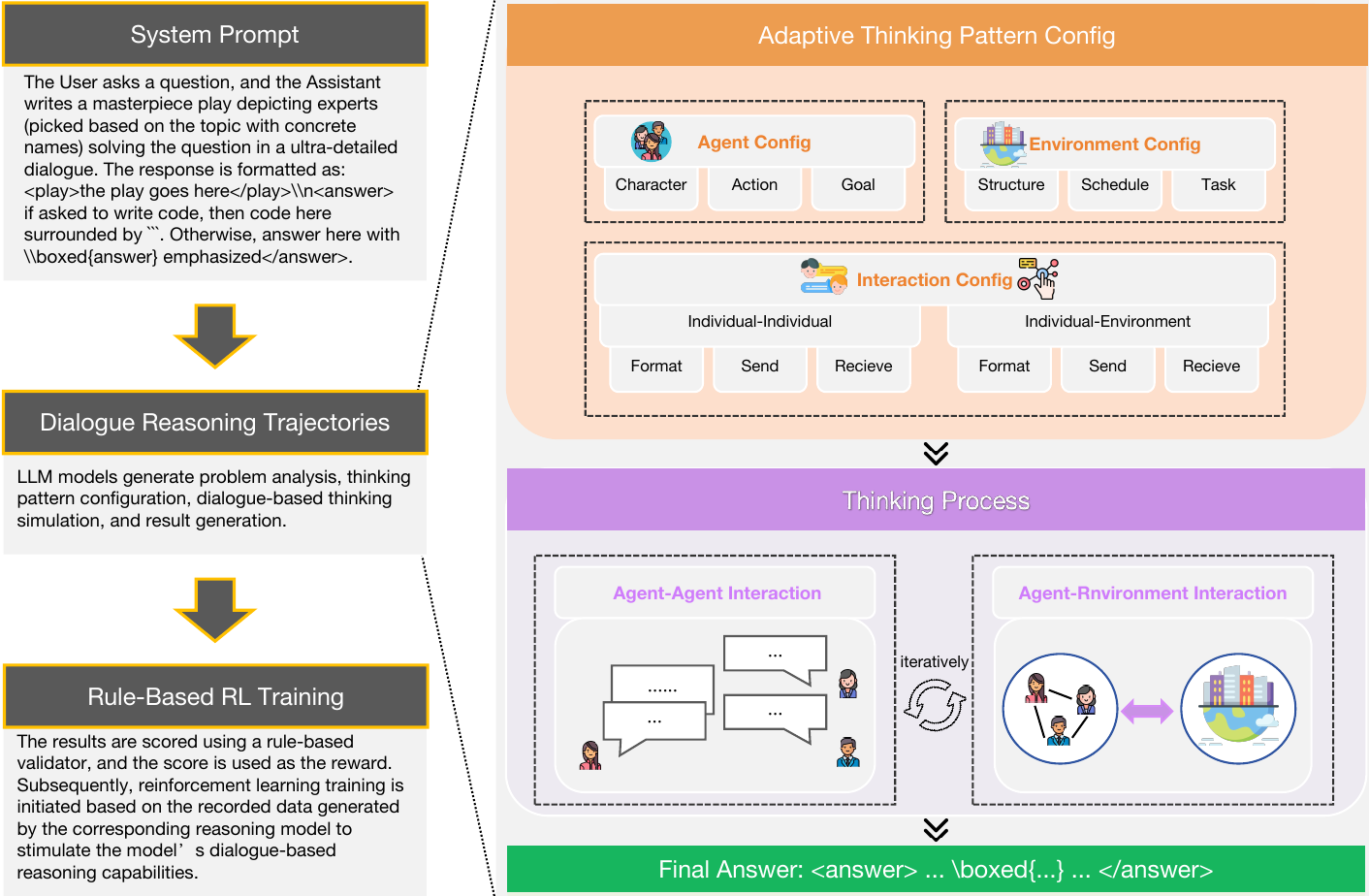}
\caption{The illustration of the dialogue-based reasoning pattern.}
\label{fig:dialogue_reason}
\end{figure}
\vspace{0pt}
\subsection{Reasoning Pattern Perspective: Dialogue-Based Reasoning}

% To enhance reasoning diversity across different problems while maintaining coherency within individual reasoning paths, we optimize reasoning from the perspective of reasoning patterns. Existing reasoning models primarily utilize a monologue-based reasoning pattern analogous to a "stream-of-consciousness," which presents notable limitations. On one hand, monologue reasoning frequently exhibits low coherency due to scattered attention, complicating the user's ability to track and understand the reasoning process. On the other hand, applying identical reasoning patterns across diverse problem types restricts the model's diversity in handling varied tasks.

To address the limitations of reasoning diversity and coherency, we propose a novel dialogue-based reasoning pattern, named \ours. As illustrated in Figure~\ref{fig:dialogue_reason}, \ours\ conceptualizes reasoning as an interactive process between distinct dialogue participants within a defined conversational setting. The figure includes a simple yet effective system prompt to stimulate dialogue-based reasoning. This prompt can be customized by specifying dialogue configurations, such as defining participants (e.g., a teacher and a student), setting the context (e.g., a math class), or choosing the dialogue format (e.g., Socratic dialogue). More systematically, inspired by agent-based simulation methodologies \cite{macal2005tutorial}, we further articulate the design space of our dialogue reasoning pattern through three key dimensions: \textbf{agents}, \textbf{environment}, and \textbf{interactions}:
\begin{itemize}[itemsep=0pt, topsep=2pt, parsep=0pt]
\item \textbf{Agent dimension}: Defines the number of reasoning agents, their designated characters, objectives, and the interests they represent;
\item \textbf{Environment dimension}: Specifies environmental functionalities, such as recording and adjusting task progression, introducing emergent events, and maintaining overall task control;
\item \textbf{Interaction dimension}:
\begin{itemize}[itemsep=0pt, topsep=2pt, parsep=0pt]
\item \textit{Agent-to-agent interactions}: Includes conflict resolution, negotiation, supplementation, and prompting among agents, represented via linguistic dialogues;
\item \textit{Agent-to-environment interactions}: Involves agents expressing requirements to the environment and the environment providing feedback to agents, thereby dynamically adjusting task goals and agent characters.
\end{itemize}
\end{itemize}

By explicitly configuring the characters and dialogue environment, the pattern encourages diverse reasoning paths tailored to different problem types. Meanwhile, the structured turn-taking and conversational boundaries inherent in dialogue promote coherency, making the reasoning process more interpretable and logically organized.

Subsequently, we draw inspiration from the latest reasoning models that leverage rule-based RL methods to continuously stimulate reasoning abilities. Accordingly, we employ a rule-based PPO approach to train LLMs to progressively develop dialogue pattern reasoning capabilities, including character configuration, dialogue simulation, and environment management.

\paragraph{Rule-based RL training to adopt the dialogue pattern.} We employ PPO~\cite{schulman2017proximal} to train models to adopt dialogue-based reasoning patterns. The actor model is selected from the Qwen series, specifically exploring both Qwen-QWQ-32B(denoted as QWQ) and the Qwen2.5-Base-32B(denoted as Base). The reward function is based on result matching. For training dataset, we utilize Open-Reasoner-Zero~(ORZ)~\cite{hu2025open}, an open-source implementation designed for large-scale reasoning-oriented RL training\footnote{\url{https://github.com/Open-Reasoner-Zero/Open-Reasoner-Zero}}, including AIME (up to 2023), OpenR1-Math-220k, Tulu3 MATH and other sources. And according to ORZ, we adopt similar experimental setups to evaluate performance on MATH-500, GPQA-Diamond, and AIME24 datasets. Our PPO optimization objective is expressed as follows:
$$
\mathbb{E}_{s,a \sim \pi_{\theta'}}\left[
  \min\left(
    \frac{\pi_\theta(a|s)}{\pi_{\theta'}(a|s)} A(s,a),\ 
    \mathrm{clip}\left(
      \frac{\pi_\theta(a|s)}{\pi_{\theta'}(a|s)},\ 1-\epsilon,\ 1+\epsilon
    \right) A(s,a)
  \right)
\right].
$$

\noindent Here, $\pi_{\theta}(a|s)$ denotes the current policy, $\pi_{\theta'}(a|s)$ denotes the old policy, and $\epsilon$ is the clipping parameter. The advantage function $A$ is calculated using Generalized Advantage Estimation (GAE):
% $$
% \hat{A}_\pi^{(\mathrm{GAE})} = \sum_{l=0}^{\infty} (\lambda \gamma)^l \delta^V_{t+l}
% $$

$$
\hat{A}_\pi^{(\mathrm{GAE})} = \sum_{l=0}^{T-t-1} \lambda^l \left( R_{t+l} + V_\phi(s_{t+l+1}) - V_\phi(s_{t+l}) \right).
$$

\noindent The value function $V_\phi$ is optimized by minimizing temporal-difference residuals. Specifically, for an individual data point, the objection function is defined as:
$$
J_V(\phi) = \frac{1}{2} ( R_t + \gamma V_\phi(s_{t+1}) - V_\phi(s_t) ).
$$
Our training discovers that both the Base model and QWQ can acquire dialogue reasoning patterns. Comparatively, \ours\ derived from the Base model exhibits dialogue reasoning that closely aligns with natural conversational structures, while \ours\ from QWQ retains a more monologue-style expression within each speak of individual characters. To ensure fair comparisons with the monologue reasoning model, our quantitative evaluations in Sec~\ref{sec:dia_results} contrast the performance of QWQ with \ours-QWQ. Additionally, we analyze the \ours-Base in Sec~\ref{sec:dia_analysis} to emphasize the distinctive characteristics of dialogue-style reasoning.

\subsection{Results}
\label{sec:dia_results}
As the number of questions combined into a single input increases, \ours\ shows stronger robustness compared to the original monologue-based approach. The comparison suggests that the dialogue-based reasoning more effectively maintains model performance under challenging conditions that demand high reasoning diversity and coherency. We present comparative experiments to support this viewpoint.

Specifically, we first compare the absolute accuracy scores of \ours-QWQ and the monologue-based reasoning model QWQ on mathematical reasoning tasks. Experimental results show that at cbK=1, QWQ often achieves higher performance—an observation consistent with prior findings that performance can degrade when switching reasoning patterns \cite{liu2025understanding}. However, as cbK increases, the task places greater demands on the model's reasoning diversity and coherency. Under these conditions, \ours\ preserves performance more effectively and ultimately surpasses QWQ in accuracy. This advantage becomes particularly evident for cbK values greater than 3. For instance, at cbK=10, on the MATH-500 dataset, QWQ remains to 89.63\% accuracy, while the dialogue model achieves 91.48\%; at cbK8, on the more challenging AIME dataset, QWQ's accuracy drops to 35.09\%, whereas the dialogue model remains higher at 50\%; similarly, at cbK8 to cbK9, on the GPQA dataset, QWQ reaches below 49.49\%, 46.57\% and 45.66\%, while the dialogue model remain performances above 52\%.

\begin{table}[htbp]
  \centering
  \caption{Comparison between the QWQ model and \ours\ derived from QWQ.}
  \label{tab:accuracy}
  \small
  \resizebox{\textwidth}{!}{%
  \begin{tabular}{l*{10}{c}}
    \toprule
          & cbK1  & cbK2  & cbK3  & cbK4  & cbK5  & cbK6  & cbK7  & cbK8  & cbK9  & cbK10 \\
    \midrule
    \multicolumn{11}{l}{\textbf{Overall}} \\
    \midrule
    QWQ            & \textbf{0.8805} & \textbf{0.8481} & 0.8371 & 0.8299 & 0.8150 & 0.7977 & 0.7860 & 0.7784 & 0.7596 & 0.7508 \\
    \ours  & 0.8685 & 0.8439 & \textbf{0.8404} & \textbf{0.8372} & \textbf{0.8260} & \textbf{0.8235} & \textbf{0.8142} & \textbf{0.8098} & \textbf{0.7925} & \textbf{0.7859} \\

    \midrule
    \multicolumn{11}{l}{\textbf{MATH-500}} \\
    \midrule
    QWQ            & \textbf{0.9762} & \textbf{0.9535} & 0.9515 & 0.9453 & 0.9385 & 0.9296 & 0.9169 & 0.9163 & 0.9016 & 0.8963 \\
    \ours  & 0.9708 & 0.9523 & \textbf{0.9516} & \textbf{0.9501} & \textbf{0.9476} & \textbf{0.9432} & \textbf{0.9375} & \textbf{0.9364} & \textbf{0.9205} & \textbf{0.9148} \\

    \midrule
    \multicolumn{11}{l}{\textbf{AIME24}} \\
    \midrule
    QWQ            & \textbf{0.7438} & 0.6458 & 0.5813 & 0.5527 & 0.4771 & 0.4542 & 0.4464 & 0.3509 & 0.3338 & 0.2667 \\
    \ours  & 0.7312 & \textbf{0.6500} & \textbf{0.6396} & \textbf{0.5996} & \textbf{0.5875} & \textbf{0.5813} & \textbf{0.5482} & \textbf{0.5000} & \textbf{0.4410} & \textbf{0.3771} \\

    \midrule
    \multicolumn{11}{l}{\textbf{GPQA-Diamond}} \\
    \midrule
    QWQ            & \textbf{0.6594} & \textbf{0.6127} & 0.5868 & 0.5806 & 0.5543 & 0.5167 & 0.5068 & 0.4949 & 0.4657 & 0.4566 \\
    \ours  & 0.6310 & 0.5994 & \textbf{0.5900} & \textbf{0.5881} & \textbf{0.5549} & \textbf{0.5578} & \textbf{0.5431} & \textbf{0.5371} & \textbf{0.5226} & \textbf{0.5224} \\
    \bottomrule
  \end{tabular}%
  }
\end{table}

\begin{figure}[H]
    \centering
    % 左边的大图
    \begin{subfigure}[b]{0.45\textwidth}
        \centering
        \begin{subfigure}[b]{\textwidth}
            \includegraphics[width=\textwidth]{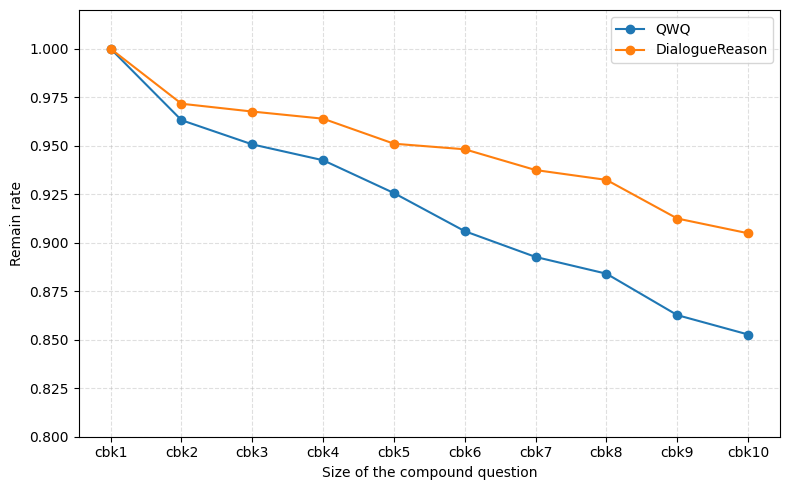}
            \caption{Overall}
            \label{fig:remain_rate_overall}
        \end{subfigure}
        
        \vspace{0.02\textwidth}
       
        \begin{subfigure}[b]{\textwidth}
            \includegraphics[width=\textwidth]{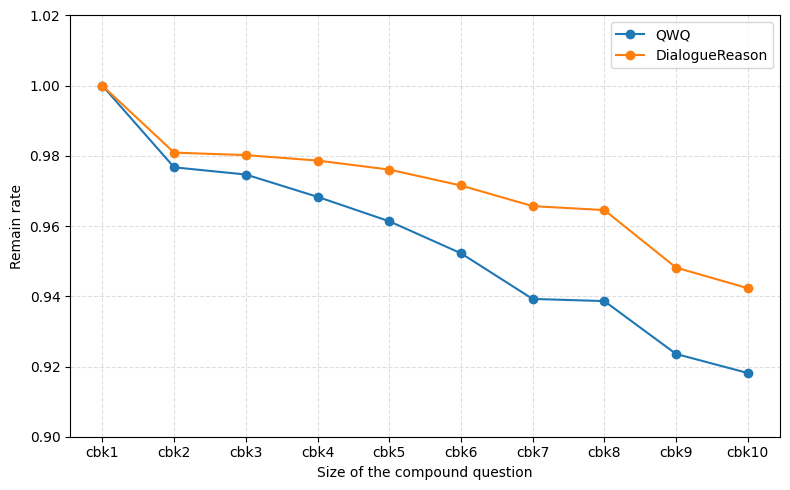}
            \caption{MATH-500}
            \label{fig:remain_rate_math}
        \end{subfigure}
    \end{subfigure}
    \hspace{0.04\textwidth} % 控制左右图之间的间距，可根据需求调整
    % 右边两张小图
    \begin{subfigure}[b]{0.45\textwidth}
        \centering
        \begin{subfigure}[b]{\textwidth}
            \includegraphics[width=\textwidth]{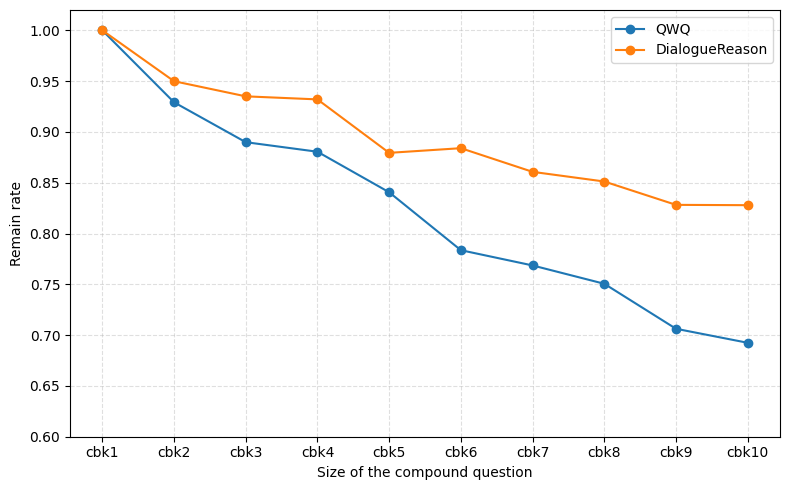}
            \caption{AIME24}
            \label{fig:remain_rate_aime}
        \end{subfigure}
        
        \vspace{0.02\textwidth}
       
        \begin{subfigure}[b]{\textwidth}
            \includegraphics[width=\textwidth]{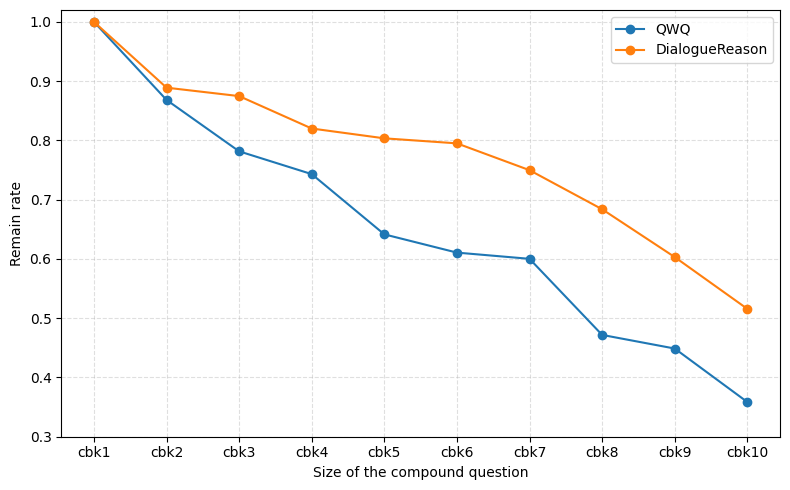}
            \caption{GPQA-Diamond}
            \label{fig:remain_rate_gpqa}
        \end{subfigure}
    \end{subfigure}
    \caption{The four subplots correspond to Overall (a), MATH‑500 (b), AIME24 (c), and GPQA-Diamond (d), respectively. The x-axis represents the compound question size from 1 to 10, while the y-axis shows the Remain Rate — the accuracy at each cbK divided by the accuracy at cbK1. A value closer to 1 indicates more stable performance.}
    \label{fig:remain_rate}
\end{figure}

Furthermore, we analyze the relative performance remaining rate, defined as the ratio of a model's accuracy at a given combination factor $k$ (cbK) to its accuracy on single-question inputs (cbK=1). We adopt this metric because compound and single questions cover identical content domains; thus, reduced performance under compound conditions highlights interference effects among reasoning processes. As illustrated in Figure~\ref{fig:remain_rate}, \ours\ (orange curves) consistently exhibits a higher remaining rate compared to QWQ (blue curve) across all datasets, further exhibiting a robust performance as the complexity of compound questions increases.

\vspace{0pt}

\subsection{Analysis}
\label{sec:dia_analysis}

\begin{figure}[htbp]
\centering
\includegraphics[width=0.95\textwidth]{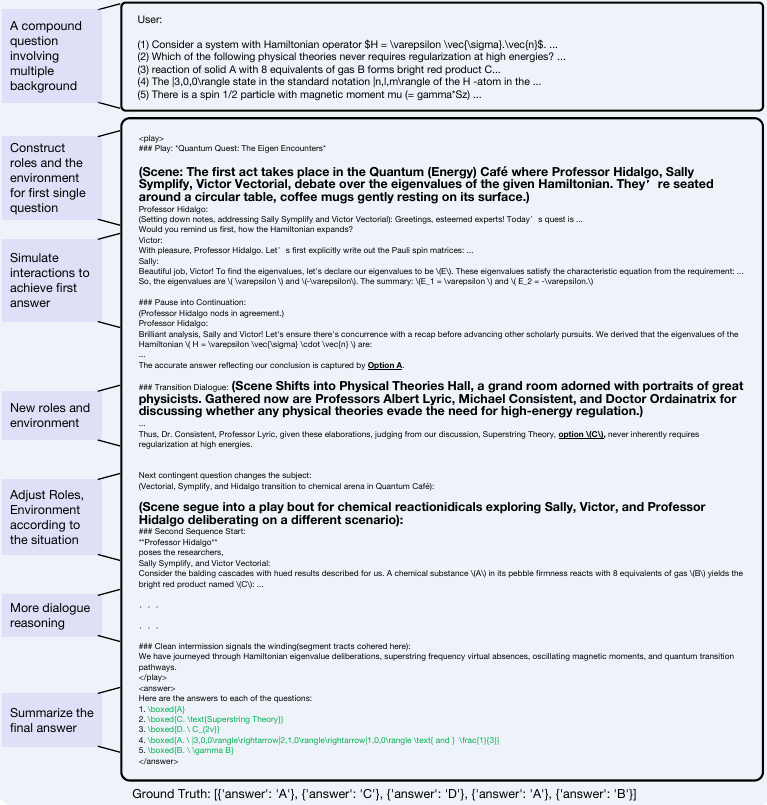}
\caption{A case study of the dialogue reasoning on a compound question from the GPQA dataset.}
\label{fig:dialogue_gpqa_case}
\end{figure}

To demonstrate the reasoning process developed by \ours\ through dialogue pattern combined with PPO, we organize and show the process in Figure~\ref{fig:dialogue_gpqa_case}. Specifically, the model first sets up a dedicated scene for each question (such as the "Quantum Café") and introduces multiple characters, each with distinct areas of expertise, within that scene. A specific problem to be solved is then posed. Through dialogue, the characters engage in interactions, progressively proposing, discussing, and refining possible solutions until they reach a consensus and summarize the final answer. When transitioning to the next question, the model constructs a new environment based on the background of the new problem (for example, shifting from the "Quantum Café" to the "Theoretical Physics Hall") and introduces a different set of characters distinct from those in the previous discussion. This approach of switching between multiple scenes and characters not only effectively prevents interference between questions and enhances the diversity of reasoning strategies, but also promotes dialogue coherency by maintaining clear character boundaries, thereby providing a more intuitive presentation of the model's complete thought process.

\subsection{Out of Distribution Question}

Given that \ours\ was trained mainly on tasks where correctness can be verified using a matching verifier, we further explore its behavior on the out-of-distribution (OOD) task to assess its generalization ability.

\begin{figure}[H]
    \centering
    % 左边
    \begin{subfigure}[b]{0.37\textwidth}
        \includegraphics[width=\textwidth]{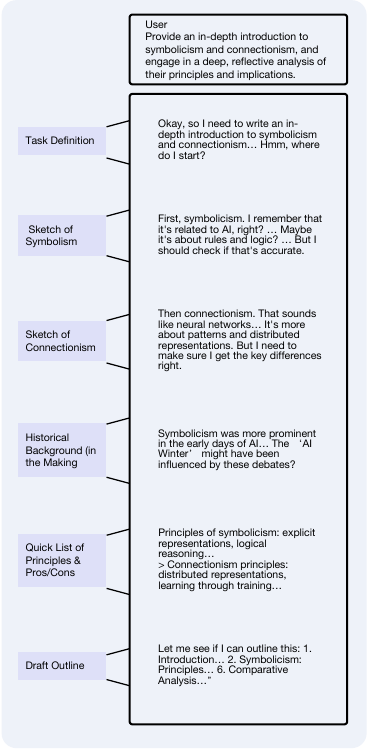} 
        \caption{Monologue}
        \label{fig:ood_cases_monologue}
    \end{subfigure}
    \hfill
    % 右边
    \begin{subfigure}[b]{0.62\textwidth}
        \includegraphics[width=\textwidth]{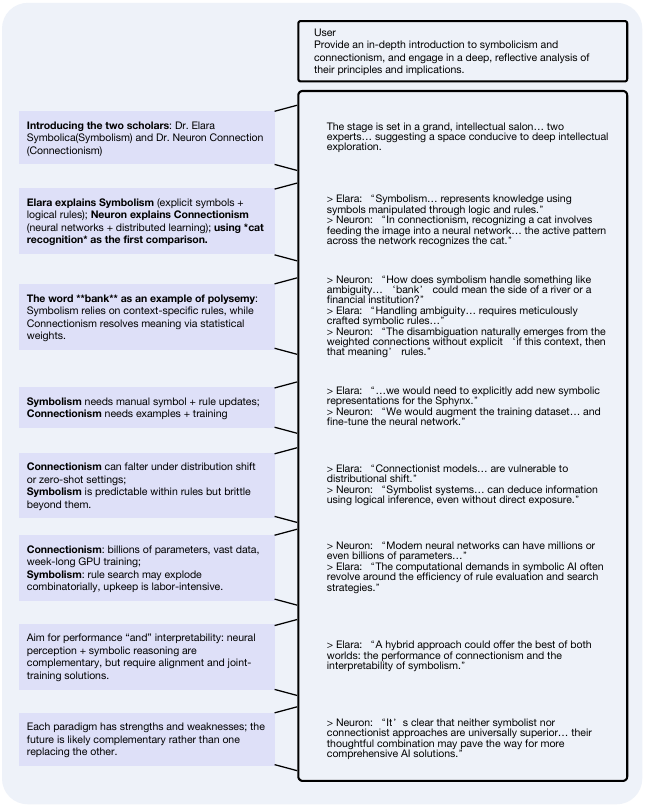}
        \caption{Dialogue}
        \label{fig:ood_cases_dialogue}
    \end{subfigure}
    \caption{The case of monologue vs. dialogue reasoning on the out-of-distribution question.}
    \label{fig:ood_cases}
\end{figure}

As illustrated in Figure~\ref{fig:ood_cases_dialogue}, when responding to an OOD question requiring different perspectives, the dialogue-based model can effectively organize suitable characters to engage in discussion. For this case, each viewpoint is assigned a specific character. In contrast, as shown in Figure~\ref{fig:ood_cases_monologue}, the monologue-based reasoning models is possibly limited by a lack of diversity, which often focuses on problem understanding and planning rather than conducting substantive reasoning. The dialogue-based approach, however, enables a more detailed and structured exploration of the question.

\section{Discussion}

This section delves into the implications and advantages of the \ours\ paradigm, shedding light on its enhanced interpretability, seamless compatibility with monologue reasoning, improved controllability and user interaction, and potential to revolutionize multi-agent training.
\subsection{Interpretability}

\ours\ appears to offer improved internal coherency. Each conversational turn tends to introduce clearer semantic boundaries, potentially helping readers recognize different stages and transitions in the reasoning process. From the perspective of users, this structure may support a high-level sense-making ability without requiring line-by-line analysis. Moreover, the inherent presence of a listener or interacting party in dialogue may naturally encourage the model to generate more understandable and structured reasoning traces, especially during extended thought chains.

\subsection{Compatibility with Monologue Reasoning}
Minsky's Society of Mind constructs a model of human intelligence as an emergent phenomenon arising from the interactions of many simple, mindless components called agents. Each agent performs a basic function, and together, their interactions form the foundation of complex mental processes—hence the name "society of mind"\cite{minsky1986society}. Building on Minsky's theory, individual reasoning can be decomposed into finer-grained cognitive agents. For instance, expressions like "Wait…" commonly found in monologue-style CoT can be attributed to a reflection agent. Similarly, we can define analogy agents, counterexample agents, and other micro-level reasoning characters. In a Dialogue-Reasoning, these agents are no longer implicit; they can be explicitly modeled to interact with one another, simulating internal deliberation in a more structured, modular form. This not only preserves compatibility with monologue reasoning but also can benefit from the enhanced diversity and coherency enabled by the dialogue structure.

\subsection{Controllability and Interaction}

Compared to monologue reasoning, dialogue provides more intuitive boundaries for user intervention or further automatic step-level verification \cite{li2023making}. Additionally, users may influence the model's reasoning process by configuring the dialogue scenario or specifying characters, offering a more accessible pathway to adjust behavior. Additionally, users could engage directly in the conversation, making fine-grained interventions at specific reasoning steps. This potential for tighter user involvement may enhance transparency and adaptability in practical settings.

\subsection{Multi-Agent Training}

Many existing multi-agent reasoning approaches mainly rely on manual design, such as predefining agent characters and constructing explicit workflows. These methods often face challenges such as unclear character boundaries and ad hoc decisions about information sharing among agents. In contrast, a dialogue-based multi-agent framework may help address some of these limitations. For example: (1) The model might automatically generate appropriate agent characters based on the input query, supporting a flexible range of configurations—from monologic solvers to tool-like subagents or diverse viewpoints; (2) It may also learn to decompose responsibilities across characters, coordinating them to solve tasks from multiple perspectives. The process of defining, segmenting, and switching characters could be integrated into an end-to-end optimization pipeline. With RL as a potential training backbone, our work could support more scalable and adaptable multi-agent reasoning. 

\section{Limitations and Future Work}

This work is primarily based on training and evaluating Qwen-series models. While the results provide promising insights into dialogue-style reasoning, several limitations remain, which point to directions for future research.

\paragraph{Task and domain scope.} The current experiments are limited to math and science reasoning tasks, specifically MATH, AIME-24, and GPQA-Diamond. While these tasks are well-suited for rigorous evaluation due to their verifiable correctness, they may not capture the full range of reasoning demands found in broader domains. We plan to expand our testing to include a wider variety of domains, such as commonsense and multi-hop QA, to better understand the generality of dialogue-based reasoning patterns.

\paragraph{Compound-QA as a training objective.} In this study, Compound-QA serves only as an evaluation tool to probe the model's robustness. However, the structure of CompoundQA, comprising multiple interdependent subproblems, may also serve as a useful training objective for RL. Incorporating compound-style questions into training data or RL episodes could help models internalize and generalize reasoning strategies across multiple steps.

\paragraph{Entangled evaluation of reasoning diversity and coherency.} Our Compound-QA evaluation is designed to simultaneously stress both diversity and coherency of reasoning. However, this conflation makes it difficult to isolate and analyze the specific weaknesses in either dimension. In future work, we plan to construct more targeted benchmarks that separately evaluate reasoning diversity (e.g., strategy switching) and coherency (e.g., consistency of intermediate steps).

\section{Related work}

\subsection{RL-Based Reasoning Model}

RL-based reasoning models have demonstrated significant potential in complex reasoning tasks. RL-based reasoning models share a common characteristic: the capacity for Long-CoT reasoning. For instance, OpenAI's o-series models~\cite{jaech2024openai} highlight the effectiveness of such methods in enhancing reasoning capabilities. Similarly, DeepSeek-R1~\cite{guo2025deepseek} utilizes rule-based matching reward functions applied to foundational or supervised fine-tuned models, combined with RL to iteratively refine and reinforce correct reasoning pathways. 

Despite the promising performance of RL-driven reasoning models, recent studies have begun to uncover critical limitations. \citet{wang2025thoughts} observe that models employing extensive reasoning frequently exhibit attention shifts, struggling to maintain sustained focus on a single logical thread. This issue results in elongated reasoning processes without corresponding improvements in accuracy, a phenomenon described as analogous to "attention-deficit hyperactivity disorder" (ADHD) in large models. Further analysis by \citet{chen2024not} highlights that models such as GPT-o1, QwQ, and DeepSeek-R1, despite excelling at mimicking complex human-like reasoning patterns (e.g., diverse approaches, chain-of-thought, and repeated verification), often become inefficient when handling simpler tasks. In these cases, the reasoning paths become repetitive, wasting computational resources without providing substantial diversity or added value. Moreover, \citet{liu2025understanding} reveals that merely incorporating extensive self-reflection into reasoning models does not significantly enhance accuracy.

Motivated by these findings, our work focuses explicitly on the dimensions of reasoning diversity and coherency. We propose and employ Compound Questions—concatenations of multiple individually solvable problems—to systematically analyze and reveal the limitations of current reasoning models when faced with multitask scenarios, thoroughly investigating their modes of failure.

\subsection{Thinking Pattern of Reasoning}

The thinking pattern underlying the reasoning process plays a critical role in determining the effectiveness of LLMs in complex reasoning tasks. Existing research in this area can be categorized into two main streams: analysis of thinking patterns and methods for improving these patterns.

In terms of analysis, \citet{liu2025understanding} highlights that the reasoning performance of the same model can vary significantly depending on the prompting templates used, especially when there is a mismatch with the model's pretraining style. Such mismatches can severely degrade model performance, requiring substantial additional training data to overcome this gap. Similar sensitivities are identified by Deepseek-R1~\cite{guo2025deepseek}, who find that few-shot prompting often negatively impacts performance, advocating instead for zero-shot prompting with clearly defined problem descriptions and output formats. Furthermore, \citet{wen2025thinkpatterns} systematically compares traditional unstructured monologue patterns with structured approaches such as self-critique, self-ask, and self-debate. Their findings suggest that unstructured monologue patterns generally perform robustly, particularly in larger models, while structured strategies are beneficial mainly for smaller models but can hinder larger model performance.

Regarding improvements, \citet{yang2025reasonflux} introduces a structured template library comprising 500 effective thinking templates, optimized through hierarchical RL. Their method progressively decomposes tasks, dynamically expands reasoning templates, and enhances hierarchical reasoning capabilities. Similarly, \citet{han2023dialcot} models reasoning as a dialogue between two distinct characters—Decomposer and Solver. Additionally, \citet{chae2023dialogue} propose a dialogue chain-of-thought approach (DOCTOR), employing alignment filtering to enhance commonsense-aware conversational abilities. Besides, CoRe \cite{zhu2023solving} adopts a dual-system approach of generation and verification, incorporating feedback loops for path-level optimization. GoT \cite{besta2024graph} extends existing prompting strategies by conceptualizing thoughts as graphs, focusing primarily on traversal and aggregation of thought structures.

Compared to these prior works, our study focuses on revealing the difficulty current reasoning models face in simultaneously achieving both diversity and coherency. From the perspective of thinking patterns, we propose a dialogue-based reasoning pattern that retains the strengths of the monologue approach. We train models to adopt this dialogue pattern via RL. Our approach leverages recent advances in rule-based RL through end-to-end training of dialogue simulations. The training enables the model to dynamically instantiate agents, configure environments, and manage interactions, resulting in more flexible and robust reasoning behaviors, particularly suited for complex mathematical and scientific tasks. Experimental results demonstrate that our \ours\ not only better balances diversity and coherency in compound question settings but also maintains more stable reasoning performance under increasing task complexity.

\section{Conclusion}
This work highlights the limitations of monologue-style reasoning in handling compound questions and introduces a dialogue-based reasoning pattern. We propose Compound-QA as an evaluation task to jointly assess reasoning diversity and coherency, and show that dialogue-style models, trained with RL, exhibit greater robustness across math and science tasks. While results are promising, future work is needed to explore broader domains, disentangle evaluation metrics, and investigate Compound-QA as a potential training objective.

\bibliography{custom}

% \appendix

% \section{Appendix}

% This is an appendix.

% \end{CJK}

\end{document}